\documentclass[runningheads]{llncs}

\usepackage[T1]{fontenc}

\usepackage{graphicx}

\usepackage{amsmath}

\usepackage{dsfont}
\usepackage[ruled,vlined]{algorithm2e}
\usepackage{lipsum}

\usepackage{float}
\usepackage[dvipsnames]{xcolor}
\usepackage[numbers]{natbib}
\usepackage[noBBpl]{mathpazo}
\usepackage{array}
\usepackage{makeidx}
\usepackage{fancyhdr}
\usepackage[hang]{caption}
\usepackage{psfrag}
\usepackage{amsmath}
\usepackage{amssymb}
\usepackage{graphics}
\usepackage{graphicx}
\usepackage{bm}
\usepackage{alltt}
\usepackage{subfigure}
\usepackage{multirow}
\usepackage{color}
\usepackage{flafter}
\usepackage{makeidx}

\usepackage{tabularx,multirow}

\usepackage{wrapfig}
\usepackage{mathpazo}
\usepackage[refpage]{nomencl} 
\usepackage{rotating}
\usepackage{psfrag}
\usepackage{listings}
\usepackage[section]{placeins}

\usepackage{lettrine}
\usepackage{mwe}

\usepackage{tablefootnote}
\usepackage {booktabs}

\usepackage[english]{babel} 
\usepackage{mathdots}

\usepackage{lscape}
\usepackage{longtable}
\usepackage{array}
\usepackage{hyperref}
\usepackage{colortbl}

\usepackage{etoolbox}

\usepackage{subcaption}

\usepackage{parskip}

\usepackage{adjustbox}

\begin{document}
\title{LLpowershap: Logistic Loss-based Automated Shapley Values Feature Selection Method}
\titlerunning{LLpowershap}

\author{Iqbal Madakkatel\inst{1,2}\orcidID{0000-0003-2339-5917} \break
Elina Hypp\"{o}nen\inst{1,2}\orcidID{0000-0003-3670-9399}} 

\authorrunning{I. Madakkatel and E. Hypp\"{o}nen}

\institute{Australian Centre for Precision Health, Unit of Clinical and Health Sciences, University of South Australia, Adelaide, SA, Australia \and
South Australian Health and Medical Research Institute (SAHMRI), Adelaide, SA, Australia\\
\email{iqbal.madakkatel@unisa.edu.au}}
\maketitle    
\begin{abstract}
Shapley values have been used extensively in machine learning, not only to explain black box machine learning models, but among other tasks, also to conduct model debugging, sensitivity and fairness analyses and to select important features for robust modelling and for further follow-up analyses. Shapley values satisfy certain axioms that promote fairness in distributing contributions of features toward prediction or reducing error, after accounting for non-linear relationships and interactions when complex machine learning models are employed. Recently, a number of feature selection methods utilising Shapley values have been introduced. Here, we present a novel feature selection method, \textit{LLpowershap}, which makes use of loss-based Shapley values to identify informative features with minimal noise among the selected sets of features. Our simulation results show that \textit{LLpowershap} not only identifies higher number of informative features but outputs fewer noise features compared to other state-of-the-art feature selection methods. Benchmarking results on four real-world datasets demonstrate higher or at par predictive performance of \textit{LLpowershap} compared to other Shapley based wrapper methods, or filter methods.

\end{abstract}
\section{Introduction}

Dimensionality reduction is an important task in machine learning and data mining, considering the benefits that it brings. Feature selection (a.k.a attribute selection), among other approaches, is an extensively used dimensionality reduction method as it helps to select relevant features and to remove irrelevant (noise) and redundant features, which generally improves model performance, reduces overfitting, and improves interpretation of models~\cite{li2017feature}. It also facilitates further follow-up analyses on powerful features identified, which require domain expertise and well-established statistical procedures~\cite{madakkatel2021combining, liu2023combining}.

In this paper, we present an enhanced version of the recently introduced SHAP value-based feature selection method, powershap~\cite{verhaeghe2022powershap}, for classification problems. Firstly, we exploit Shapley values~\cite{shapley1953value} (hereafter, called as LogisticLossSHAP) distributing the mismatch between prediction and the truth calculated by the logistic loss function among the input features. Secondly, we make certain modifications in calculating p-values for identifying important features. Finally, we make further modifications in statistical power calculations for automating the feature selection method and thus removing the need to provide the number of iterations that the method needs to run. We call our method \textit{LLpowershap} (Logistic Loss-based powershap).

In the following section, we illustrate the connection and benefits of using \textit{LLpowershap} compared to the previous methods. In Section~\ref{LLpowershap}, we explain in detail the changes that we make to improve upon powershap~\cite{verhaeghe2022powershap}. By following the testing strategy described in~\cite{verhaeghe2022powershap}, the performance of \textit{LLpowershap} is then compared against five other state-of-the-art Shapley value-based methods and two filter methods using both simulation and real-world datasets (Sections~\ref{experiments} and~\ref{results}). Section~\ref{discussion} discusses the results, and the conclusion is provided in Section~\ref{conclusion}.

\section{Background}\label{relatedwork}

Post-hoc explanations of black-box machine learning models are becoming increasingly popular, with SHAP~\cite{lundberg2017unified}, an additive feature attribution method, becoming a common method in providing local explanations with good local fidelity. SHAP values are based on Shapley values, a unique solution concept used in game theory that deals with how fairly (mathematically) the payoffs from a cooperative game can be distributed to the game players, satisfying certain axioms, namely, dummy/null player, efficiency/full allocation, symmetry/fairness, and linearity. Shapley value for a player is defined as the average marginal contribution of the player, considering all possible ways that the player can be part of. It has been demonstrated that Shapley value-based methods can provide different attributions to features for the same input even when the values are computed exactly (as opposed to approximation), owing to different cooperative games formulated by the methods~\cite{merrick2020explanation}. Marginal (also called interventional or true to the model) Shapley values use marginal distributions to simulate missingness of features as required in the calculation of Shapley values. Conditional (also called true to the data) Shapley values, on the other hand, use a conditional distribution to simulate missingness of features. Conditional Shapley values tend to spread credit between correlated features, whereas \sloppy{marginal} Shapley values produce attributions that are a description of the functional form of the model~\cite{chen2023algorithms}. Conditional and marginal approaches are by far the most common feature removal approaches in practice~\cite{chen2023algorithms}.

TreeSHAP~\cite{lundberg2020local}, is a widely used algorithm to calculate/estimate Shapley values for tree-based models (decisions trees, tree ensembles such as random forests and gradient boosting), leveraging the structure of the tree(s) to reduce time complexity from exponential to polynomial time. TreeSHAP is not model agnostic because it utilises internal details of decision trees such as the values at the leaves and the splitting proportions at internal nodes. It has a conditional version, known as Path-dependent TreeSHAP (which estimates Shapley values, biased but variance free) and a marginal version, Interventional TreeSHAP (which calculates exact Shapley values, unbiased and variance free)~\cite{chen2023algorithms}. Interventional TreeSHAP calculates exact marginal Shapley values in time linear in the size of the model and number of samples in the background dataset used for simulating feature removal~\cite{lundberg2020local}, with the advantage of having no burden of checking convergence and also having no noise in the estimates~\cite{lundberg2020local}. Having no sampling variability when applied to models with many input features is an advantage of using Interventional TreeSHAP. We use Interventional TreeSHAP in \textit{LLpowershap} by setting \texttt{feature\_pertubation=``interventional''} in creating explainer objects. Interventional TreeSHAP can be slower than Path-dependent TreeSHAP if the background dataset is large as the complexity scales with the size of the background dataset. However, model agnostic explainers such as KernelSHAP~\cite{lundberg2017unified} and SAGE~\cite{covert2020understanding} can be much slower compared to Interventional TreeSHAP.

Rank-based feature importance feature selection method and other Shapley value-based methods need global feature importance, rather than local Shapley values. One common way of finding global feature importance is by averaging absolute local Shapley values, so that positive and negative values do not cancel out.  However, there are methods that estimate/calculate global feature importance by assessing features effect on loss~\cite{covert2020understanding, tripathi2020interpretable}. For example, the work cited in~\cite{tripathi2020interpretable} follows the strategy of using linear models with hinge loss-based characteristic function and relate a feature's contribution to Shapley value-based error apportioning (SVEA) of total training error, enabling to use a zero-based threshold to identify important subset of features. 

To explain interventional Shapley values for models, assume we have a trained model $f$ to predict $Y$ given an input $X$, consisting of individual features \sloppy{$(X_1, X_2, \ldots, X_m)$}. With the subset of given features as $X_S \equiv \{X_i | i \in S\}$ and the subset of missing features to be simulated as $X_{\bar{S}} \equiv \{X_i | i \notin S\}$, the prediction $f(x_S, X_{\bar{S}})$ in interventional Shapley values can be found as

\begin{equation}
f(x_S, X_{\bar{S}}) = \mathbb{E}_{X_{\overline{S}} \sim \mathcal{P}(X)} [ f(X_S, X_{\bar{S}} | X_S = x_S)].
\end{equation}

Instead of explaining the prediction made by the model $f$, we can marginally explain the logistic loss $\ell$. Then the characteristic function for the cooperative game will be

\begin{equation}
v_{f,\ell,X,Y}(S) =  \mathbb{E}_{X_{\overline{S}} \sim \mathcal{P}(X)} [ \ell(f(X_S, X_{\bar{S}} | X_S = x_S), Y)] -  \mathbb{E} [ \ell(f(X), Y)].
\end{equation}

Due to the popularity, accessibility and the strength of the SHAP algorithm, in addition to explaining models and other tasks, they have been used in the development of new feature selection methods. Rank-based method using either a rank cut-off or Shapley value cut-off to determine important subset of features is straightforward and is commonly used. Other methods using SHAP values include BorutaShap~\cite{keany2020borutashap}, shapicant~\cite{calzolari2020shapicant} and the recent powershap~\cite{verhaeghe2022powershap}. In BorutaShap (based on Boruta algorithm), utilising TreeSHAP (and hence only works for tree based models), every input feature is randomly shuffled to create shadow features. It works on the idea that a feature is important if it is only better than the best performing shadow feature. The algorithm is repeated for a number of iterations for statistical interpretation for selecting important features based on a cut-off value for p-values.
Instead of permuting input features, in shapicant, labels are permuted. First, the true SHAP values are calculated using the actual labels and then for a number of iterations, the labels are randomly shuffled, models are trained and null SHAP values are calculated. Given the true SHAP values and the sets of null SHAP values, statistical interpretation can be realised by non-parametrically estimating p-values and thus enabling the selection of important features using a cut-off value for the p-values obtained. Shapicant specifically uses both the mean of the negative and positive Shapley values.

Powershap algorithm has two parts: the \textit{Explain} component and the \textit{Core} component. In the first part, when training a model, a uniform random noise feature is added to the model input and the importance of features (including the random noise feature) is calculated on the validation set (which is used during the training process for early stopping). This process is repeated for a number of iterations, each time the random noise is generated by setting a different random seed for random number generation as well as for splitting the training samples into training and validation sets. Shapley values are aggregated by calculating mean absolute Shapley values of individual samples.  In the \textit{Core} part of the algorithm, p-value for each feature is calculated using a percentile formulae ranking the position of mean (for all the iterations) Shapley value of a feature  among the Shapley values of the random noise feature obtained from each iteration. Furthermore, it has an automatic mode which uses statistical power calculations by heuristally assuming t-distributions for noise and feature Shapley values to determine the number of iterations required to reach the required power level  ($\beta$) for a given p-value cut-off ($\alpha$). This eliminates the need to provide the hyper-parameter for number of iterations that the method needs to iterate before calculating p-values.  It is noteworthy that BorutaShap and powershap also calculate global feature importance by averaging ($\mu$) the absolute local Shapley values of each feature, while shapicant uses both the mean of positive and negative Shapley values. 

\subsection{Other feature selection methods}

Generally, feature selection approaches are classified into three broad categories, namely, filter, wrapper and embedded methods. On most occasions, the aim of feature selection is to find the optimal set of features with maximum accuracy~\cite{kohavi1997wrappers}. Similar to powershap, BorutaShap and shapicant, our method belongs to the category of wrapper methods.

Filter methods have several advantages as (a) no models need to be trained, (b) they are easy to implement and fast to execute, and (c) they can handle higher number of features. A recent study shows that there is no one criterion better than others and that which criterion should be used depends upon the specific problem in hand~\cite{bommert2020benchmark}. These approaches have limitations. In addition to imposing assumptions on the data (for example, F test assumes a linear relationship between the feature and the outcome variable), these methods do not consider interactions among features and often require a cut-off point specification or hyper-parameter tuning~\cite{kumari2011filter}.  Wrapper methods invariably involve training models which can capture non-linear relationships and complex interactions. Greedy sequential search such as backward and forward selection and recursive feature elimination are examples of wrapper methods. Wrapper methods can be slow and computationally inefficient, however they generally return fewer but more qualitative features  accounting for non-linearity and interactions. In embedded feature selection methods, feature selection is part of training the model itself. For example, fitted trees in random forest models or gradient boosting models can be used to calculate feature importance (e.g., \textit{PredictionValuesChange} in CatBoost~\cite{dorogush2018catboost} models) facilitating subsequent feature selection.  Although these approaches can capture non-linearity and interactions (for example, feature importances calculated in tree models), these methods are highly dependent upon the the algorithms used for creating the models.  Often, they also lack axiomatic properties (efficiency, symmetry, dummy player, and additivity) of Shapley values. Embedded models also include penalised linear models for feature selection, such as LASSO regression, which requires creating additional features for capturing non-linearity and interactions.

\section{LLpowershap}\label{LLpowershap}

Like powershap, \textit{LLpowershap} builds on the idea that a known random feature should have on average lower impact on predictions than an informative feature, but extends the idea in two ways. Firstly, we utilise loss-based Shapley (LogisticLossSHAP) values as it considers the loss instead of model prediction, which can have an impact when a model does not predict the output well.  Moreover, we calculate LogisticLossSHAP values on truly unseen data rather than on training/validation sets. The reason for this change is that our observations show that LogisticLossSHAP values calculated on test sets have lower Shapley values for noise features compared to the values calculated on the training sets, that is, the strength of the noise features is further attenuated by calculating Shapley value on truly unseen data. On our simulation data, this attenuation for 100 iterations ranged roughly from 1.5 to 4 times, whereas we did not observe such attenuation in informative features. Secondly, substantiated by the observations (for example, from 0 noise features to 15 noise features in the output) that using a single noise feature results in greater number of noise output when LogisticLossSHAP values are used, a new powerful set of LogisticLossSHAP values is created by taking the maximum of different noises of different standard distributions in \emph{each iteration}.  As expected, on the simulation datasets, statistical analyses show that this newly created set of LogisticLossSHAP values has higher influence than the individual noises. We insert five different noise features (uniform, normal, exponential, logistic and Cauchy standard distributions). Using this set of values is somewhat similar to the strategy followed in BorutaShap, but in BorutaShap the shadow features are created for every input feature by randomly shuffling the input features and picking up the best performing shadow feature. Creating shadow features for all features can result in time and space complexity issues when training data is very large. In principle, it may also be possible to employ noises of the same distribution type such as uniform distribution in \textit{LLpowershap}, but the current strategy is more aligned with the shadow feature strategy of BorutaShap, that the real features often exhibit more specific distributions. For example, in large datasets, income of participants often exhibits log-normal characteristic with decrease in frequencies as the income levels go up, whereas measures such as height, weight and blood pressure of participants usually follow a normal distribution.

Positive and negative Shapley values of a feature for prediction indicate the direction and magnitude of impact of the feature to reach the predicted value from a global baseline value. Both higher positive and negative Shapley values are valuable for prediction. Unfortunately, tree-based models sometimes use noise features in creating decisions trees. Also, tree explainers sometimes provide non-zero Shapley values to noise features~\cite{linardatos2020explainable}, making it less attractive to use zero-based threshold on mean absolute local Shapley values to select informative features. In the case of Shapley values calculated for explaining logistic loss, positive and negative values have different meanings. Positive Shapley values increase the loss and negative values reduce the loss. We negate Shapley values for easier understanding. Again, due to the issues aforementioned, zero-based threshold is not ideal. We do not take the absolute of LogisticLossSHAP values as it is not appropriate. 

In Algorithm~\ref{alg1}, we show how Shapley values are calculated in \textit{LLpowershap}, which is a modification of the algorithm proposed by \citet{verhaeghe2022powershap}. The key differences are (a) using LogisticLossSHAP instead of SHAP values, (b) splitting the whole training samples at the proportion of 0.7 and 0.1 for training and validation and 0.2 for the test set (to calculate LogisticLossSHAP values on truly unseen data) and (c) employing different noise features and taking the maximum in each iteration. 

\begin{algorithm}[h]
    \SetAlgoLined

    \SetKwFunction{FMain}{Explain}
    \SetKwProg{Fn}{Function}{:}{}
    \Fn{\FMain{$I \leftarrow Iterations$, $M \leftarrow Model$, $(\mathbf{X}^{n \times m}, \mathbf{Y}^{n}) \leftarrow Data$ , $rs \leftarrow Random\ seed$}}{\vspace{0.2cm}
    	\tcp*[h]{This algorithm is a modification of Algorithm 1 in~\cite{verhaeghe2022powershap}} \par\vspace{2mm}
        \bf{LLpowershap}$_{\textnormal{{values}}} \leftarrow \textnormal{{size}}[I, m+5]$ \par\vspace{1mm}
        \For{$i \leftarrow 1,2,\ldots,I$}{ \vspace{1mm} \par\vspace{1mm}
            $RS \leftarrow i + rs$ \par\vspace{1mm}
            $\mathbf{O}^{n}_{rand\_unif} \leftarrow \textnormal{RandomUniform}(RS)  \textnormal{{ size }} n$  \par\vspace{1mm}
	    $\mathbf{O}^{n}_{rand\_norm} \leftarrow \textnormal{RandomNormal}(RS)  \textnormal{{ size }} n$  \par\vspace{1mm}            
	    $\mathbf{O}^{n}_{rand\_logi} \leftarrow \textnormal{RandomLogisitc}(RS)  \textnormal{{ size }} n$  \par\vspace{1mm}    
	    $\mathbf{O}^{n}_{rand\_expo} \leftarrow \textnormal{RandomExponential}(RS)  \textnormal{{ size }} n$  \par\vspace{1mm}    	
	    $\mathbf{O}^{n}_{rand\_cauchy} \leftarrow \textnormal{RandomCauchy}(RS)  \textnormal{{ size }} n$  \par\vspace{2mm}

	    $ \mathbf{X}^{n \times m+1..m+5} \leftarrow  \mathbf{X}^{n \times m} \cup  \mathbf{O}^{n}_{rand\_unif} \cup \mathbf{O}^{n}_{rand\_norm} \cup \mathbf{O}^{n}_{rand\_logi} \cup \mathbf{O}^{n}_{rand\_expo} \cup \mathbf{O}^{n}_{rand\_cauchy}$   \par\vspace{2mm}

	    $ (\mathbf{X}^{0.7n \times m+5}_{train}, \mathbf{Y}^{0.7n}_{train}),  (\mathbf{X}^{0.1n \times m+5}_{val}, \mathbf{Y}^{0.1n}_{val}), (\mathbf{X}^{0.2n \times m+5}_{test}, \mathbf{Y}^{0.2n}_{test}) \leftarrow \textnormal{split} \ (\mathbf{X}, \mathbf{Y}$)  \par\vspace{2mm} 	    

            $M \leftarrow Fit\ M((\mathbf{X}_{train}, \mathbf{Y}_{train}), (\mathbf{X}_{val}, \mathbf{Y}_{val}))$ \par\vspace{1mm}
            $\mathbf{s}_{values} \leftarrow LogisticLossSHAP(M, (\mathbf{X}_{test}, \mathbf{Y}_{test}))$ \par\vspace{1mm}
            \For{$j \leftarrow 1,2,\ldots,m + 5$}{ \par\vspace{1mm}
                \bf{LLpowershap}$_{\textnormal{{values}}}[i][j] \leftarrow (\mu(\mathbf{s}_{values}[...\ ][j]))$ \par\vspace{1mm}
            }
        }
        return  \bf{LLpowershap}$_{\textnormal{values}}$
    }
    \caption{\textit{LLpowershap} Explain algorithm}
    \label{alg1}
\end{algorithm}

Powershap uses a percentile formula to calculate p-value empirically, ranking the mean Shapley value for a feature within the Shapley values for the random uniform feature. In contrast, for each iteration we consider only the noise feature that has had the highest influence. To empirically calculate p-value we consider the distribution of Shapley values of both the highest noise and the feature as non-parametric and employ the Mann-Whitney U test. The null hypothesis here is that there is no difference between the distributions and the alternative hypothesis is that Shapley value distribution of the feature is greater than that of the noise. This results in getting a p-value for each feature.   Then using a user supplied p-value cut-off $\alpha$, we can find the set of informative features. We use \texttt{scipy.stats.mannwhitneyu}~\cite{virtanen2020scipy} to calculate p-values.  Algorithm~\ref{alg2} shows the modifications to calculate p-values.

\begin{algorithm}[h]
    \SetAlgoLined
    
    \SetKwFunction{FMain}{LLpowershap}
    \SetKwProg{Fn}{Function}{:}{}
    \Fn{\FMain{$I \leftarrow Iterations$, $M \leftarrow Model$, $\mathbf{F}_{set} \leftarrow F_{1}, ..., F_{m}$, $(\mathbf{X}^{n \times m}, \mathbf{Y}^{n}) \leftarrow Data$, $\alpha \leftarrow \ p$-\textit{value} cut-off}}{\vspace{0.2cm}
    	\tcp*[h]{This algorithm is a modification of Algorithm 2 in~\cite{verhaeghe2022powershap}} \par\vspace{2mm}
        \bf{LLpowershap}$_{\textnormal{{values}}} \leftarrow  $\bf{Explain}($I, M, \mathbf{X}, \mathbf{Y} $\textnormal{)}\par\vspace{1mm}
        $\mathbf{S}_{random} \leftarrow$ size[$I, 1]$\par\vspace{1mm}
        \For{$i \leftarrow 1,2,\ldots, I$}{\par\vspace{1mm}
        	$\mathbf{S}_{random}[i] \leftarrow$ \textnormal{Max(}\bf{LLpowershap}$_{\textnormal{values}}[i][m+1,\ldots, m+5]$\textnormal{)}\par\vspace{1mm}
 	}
        \For{$j \leftarrow 1,2,\ldots, m$}{\par\vspace{1mm}
        	
        	$\mathbf{P}_{j} \leftarrow$ \textnormal{MannWhiteneyU(}\bf{LLpowershap}$_{\textnormal{values}}$\textnormal{[\ldots][j],} $\mathbf{S}_{random}$\textnormal{)}
 	}	
        \Return{$\{F_i\ |\  \forall\  i : P[i] < \alpha\}$}
    }
    \caption{\textit{LLpowershap} Core algorithm}
    \label{alg2}
\end{algorithm}        
    
\subsubsection{Automatic Mode}

We utilise the powerful automatic mode designed in~\cite{verhaeghe2022powershap} with the following changes. We set the initial number of iterations to 20 instead of 10 iterations which helps to estimate more accurate p-values. The other change is in the calculation of the effect size. We test the equality of variance using Levene’s test and if there is no difference in variances to the user provided $\alpha$ value, we use pooled standard deviation (equation~\ref{pooled} and~\ref{effpooled}). Where variances are not equal, we use Glass’s delta (equation~\ref{glass}, using the standard deviations of Shapley values of features) in estimating the effect size. We find that, generally, for the LogisticLossSHAP values, the injected noise features have much smaller means (roughly 50 to 1,000 times less) and standard deviations (5 to 100 times less) compared to informative features in our simulation data. We assume t-distributions (our observations on the simulation data show this is a reasonable assumption) both for the noise and informative features for empirical power calculations. Algorithm~\ref{alg3} shows the power analysis. We do not make any changes to the following aspects of the automatic mode of powershap. That is, the default $\alpha$ is set to 0.01 and the required power to 0.99. If the number of required iterations estimated using the power calculation exceeds 10, additional 10 iterations are added and p-values and required iterations are again calculated. By default, this is repeated three times to avoid a possible infinite calculation. We provide source files that can be used to replace certain existing files in powershap~\cite{verhaeghe2022powershap} to make use of \textit{LLpowershap}.
\begin{equation}
s_p = \sqrt{\frac{s_1^2 + s_2^2}{2}}\label{pooled}
\end{equation}
\begin{equation}
effect\_size = \frac{\mu(s_1) - \mu(s_2)}{s_p}\label{effpooled}
\end{equation}
\begin{equation}
effect\_size = \frac{\mu(s_1) -\mu(s_2)}{s_2}\label{glass}
\end{equation}
\begin{algorithm}[h]
    \SetAlgoLined
    \SetKwFunction{FMain}{Analysis}
    \SetKwProg{Fn}{Function}{:}{}
    \Fn{\FMain{$\alpha \leftarrow \ p$-\textit{value} cut-off, $\beta \leftarrow required\ power$, \bf{LLpowershap}$_{\textnormal{{values}}}$}}{\vspace{0.2cm}
    	\tcp*[h]{This algorithm is a modification of Algorithm 3 in~\cite{verhaeghe2022powershap}} \par\vspace{2mm}

        $\mathbf{S}_{random} \leftarrow$ size[$I, 1]$\par\vspace{1mm}
        \For{$i \leftarrow 1,2,\ldots, I$}{\par\vspace{1mm}
        	$\mathbf{S}_{random}[i] \leftarrow$ \textnormal{Max(}\bf{LLpowershap}$_{\textnormal{values}}[i][m+1,\ldots, m+5]$\textnormal{)}\par\vspace{1mm}
 	}
 	    	
	$\mathbf{P} \leftarrow $ size [$m$]\par\vspace{1mm}
	
	$\mathbf{N}_{required} \leftarrow $ size [$m$]\par\vspace{1mm}

        \For{$j \leftarrow 1,2,\ldots, m$}{\par\vspace{1mm}
        
         	$\mathbf{S}_{j} \leftarrow \bf{LLpowershap}_{\textnormal{values}}\textnormal{[\ldots][j]}$\par\vspace{1mm}
        	
        	$\mathbf{P}_{j} \leftarrow$ \textnormal{ManWhiteneyU(}$\mathbf{S}_{j}, \mathbf{S}_{random}$\textnormal{)} \par\vspace{1mm}
        	
        	$P_{v} \leftarrow $\textnormal{LeveneVarianceTest(}$\mathbf{S}_{j}, \mathbf{S}_{random}$\textnormal{)} \par\vspace{1mm}
        	
        	\If{$P_{v} < \alpha$}{\par\vspace{1mm}
        	
        		\textit{effect\_size} $\leftarrow$ \textnormal{EffectSizeGlassDelta(}$\mathbf{S}_{j}, \mathbf{S}_{random}$\textnormal{)} \par\vspace{1mm}
        	
        	}
        	\Else {\par\vspace{1mm}
        	
        		\textit{effect\_size} $\leftarrow$ \textnormal{EffectSizeCohensD(}$\mathbf{S}_{j}, \mathbf{S}_{random}$\textnormal{)} \par\vspace{1mm}        	
        	}
        	
        	$\mathbf{N}_{\text{{required}}}[j] \gets \text{{SolvetTTestPower}}(\text{{effect\_size}} , \alpha , \beta)$\par\vspace{1mm}
        	
 	}	
        \Return{$\mathbf{P}, \mathbf{N}_{\text{{required}}}$}
    }
    \caption{\textit{LLpowershap} analysis function}
    \label{alg3}
\end{algorithm}    

\section{Experiments}\label{experiments}

\subsection{Feature Selection Methods}

We repeat the experiments conducted in~\cite{verhaeghe2022powershap} with some additions. We include \textit{LLpowershap} and a rank-based feature selection method that heuristically considers the top 3\% of features (when there are more than 100 features to select from) as the relevant features, as used in comparing different feature selection methods on different synthetic datasets in~\cite{bolon2015feature}. We additionally include a very large real-world biomedical dataset consisting of more than 450,000 samples and over 2,800 features. We do not include forward selection due to the time complexity issues with forward selection as noted in~\cite{verhaeghe2022powershap}. To allow direct comparability, we chose to employ the same simulation and real-world datasets as in~\cite{verhaeghe2022powershap} and also used the same random seeds to generate and split the datasets. In simulation data, we additionally include the case of only 3\% of the total number of features being informative. We also test the methods with 10,000 samples in the simulation data. Two filter methods (providing p-values) selected for comparison are chi squared and F-test feature selection, available from the \texttt{sklearn} library~\cite{pedregosa2011scikit}. Input values are shifted to positive range by adding the absolute of the minimum value for chi-squared test as the test works only with positive values.  Chi-square test returns low p-value when a feature is not independent of the outcome variable. F-test fits univariate linear regression models and reports F-score and p-value with the null hypothesis that the feature does not differentiate between the two classes.  For all Shapley value based method we used an XGBoost~\cite{chen2016xgboost} gradient boosting tree-based method with 250 estimators with overfitting detection enabled (\texttt{early\_stopping\_rounds} = 25). We used an XGBoost model instead of a CatBoost model (as done in~\cite{verhaeghe2022powershap}) as CatBoost lacked Interventional TreeSHAP for logistic loss. This way we reduced/nullified algorithm specific differences (for example, CatBoost has ordered boosting and dynamic selection of learning rate based on dataset and number of estimators specified compared to XGBoost). XGBoost also has the advantage of choosing only one feature from a set of perfectly correlated features to construct decision trees and thereby rendering Shapley value of zero to those features not used. CatBoost uses all of those perfectly correlated features resulting in distributing feature importance among the correlated features. All Shapley value based methods are tested in their default mode. Both powershap and \textit{LLpowershap} supported early stopping with a validation set and hence a validation set was provided. BorutaShap and shapicant in default mode did not support early stopping.  In the experiments we were primarily interested in the returned informative features, noise features in the output and the performance of models using only the selected features. The code to reproduce the results can be found at \url{https://github.com/madakkmi/LLpowershap}.

\subsection{Simulation Data}

All the methods (except the \emph{top 3\%}) are tested on simulation data containing 5,000 and 10,000 samples, created using \texttt{make\_classification} of \texttt{sklearn}, with the hypercube parameter set to \texttt{True}. When hypercube parameter is enabled, samples are drawn in a hypercube-manner.  The simulations are run for 20, 100, 250 and 500 features with the percentage of informative features varying as 3\% (one feature if 3\% is less than one), 10\%, 33\%, 50\% and 90\%, enabling a comprehensive evaluation of the methods under different varying conditions. We repeated the experiments five times using different random seeds in invoking the \texttt{make\_classification} function. We set the number of repeat and redundant (linear combination of informative features) features as zero. The methods tested were \textit{LLpowershap}, powershap, BorutaShap, shapicant, chi-squared and F-test.

\subsection{Benchmark Datasets}

We evaluate the different methods in their default configuration on three publicly available high dimensional classification datasets, namely, Madelon~\cite{guyon2004result}, Gina priori~\cite{OpenMLD1042}, the Scene dataset~\cite{OpenMLD312}, and the UK Biobank~\cite{sudlow2015uk} data prepared for predicting cancer incidences.  Madelon is a multivariate and highly non-linear dataset, whereas Gina priori is a digit recognition dataset and Scene is a scene recognition dataset. The UK Biobank data was prepared for identifying risk factors of cancer incidence in over 400,000 UK Biobank participants after their baseline visit for a follow-up period of 10 years~\cite{madakkatel2023hypothesis}. 

 The characteristics of these datasets are shown in Table~\ref{charact}. For the Scene dataset, we used the label ``Urban” for the experiments.
The datasets are split into training and test sets at the ratio of 75:25. All methods are evaluated using 10-fold cross validation using two other independent state-of-the-art gradient boosting decision trees algorithms, CatBoost and LightGBM~\cite{ke2017lightgbm} in their default mode. Further performance was assessed on the test set using 1,000 bootstrapped datasets. The models were evaluated using the area under the receiver-operating characteristics (AUROC) metric.

\begin{table}[h]
\centering
\caption{Characteristics of the benchmarking datasets.} \smallbreak
\begin{tabular}{ccrrr}
\hline
\textbf{Dataset} & \textbf{Source} & \textbf{\# features} & \textbf{train size} & \textbf{test size} \\
\hline
Madelon  & OpenML & 500 & 1,950 & 650 \\

Gina priori  & OpenML & 784 & 2,601 & 867 \\

Scene & OpenML & 294 & 1,805 & 867 \\
UKB Cancer & UK Biobank & 2,828 & 344,376 & 114,793 \\
\hline
\end{tabular}
\label{charact}
\end{table}

\section{Results}\label{results}

\subsection{Simulation Data}

\begin{figure*}[t]
  \centering
  \rotatebox{90}{\includegraphics[width=\textheight]{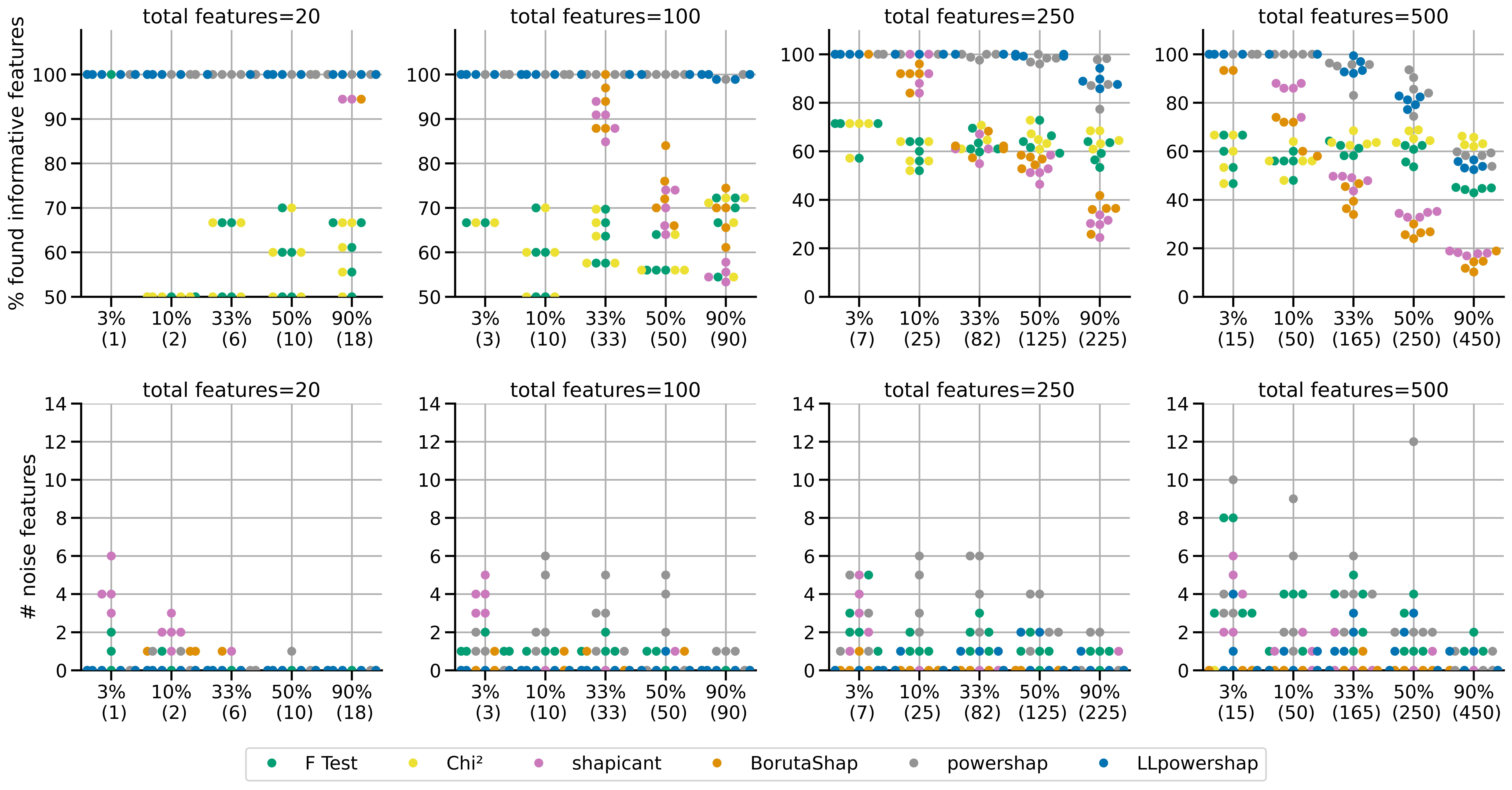}}
  \caption{Simulation benchmark results on the datasets created using the \texttt{make\_classification} of \texttt{sklearn} for 5,000 samples with five different random seeds. X-axis labels show the percentages of informative features in the datasets with the counts in brackets. Each dot of a particular colour represents result from one simulation (for example, using random seed 0) for a particular method. }
  \label{fig1}
\end{figure*}

Table~\ref{make-classi-table1} shows the results of simulation in terms of the percentage of informative features found for the datasets with sample size 5,000 and 10,000, whereas Table~\ref{make-classi-table2} shows the results in terms of number of selected noise features. For datasets with 5,000 samples, \textit{LLpowershap} returns maximum number of informative features in all cases except when there are 500 features and the percentage of informative features is 50\% and 90\%. \textit{LLpowershap} and powershap have similar performance in terms of percentage of informative features found but \textit{LLpowershap} has lower number of noise features (in terms of mean and standard deviation) in the output. Among filter methods, chi-squared method is better than F-test with higher number of informative features found, with no noise in the output. BorutaShap and shapicant have similar performance except for 100 features with 90\% of informative features.  Figure~\ref{fig1} shows the results of simulation for each seed with the first row for the percentage of informative features found and the second row for number of selected noise features for the dataset with 5,000 samples. The results show higher noise output for powershap compared to \textit{LLpowershap}. For 10,000 samples, \textit{LLpowershap} and powershap have comparable performance in terms of percentage of informative features found but in most settings, \textit{LLpowershap} outputs lower number of noise features. It returns nil or nearly nil noises for features 20, 100 and 250, and lower number of noises compared to powershap for 500 features. We observe that filter methods, chi-squared and F-test do not improve performance with the addition of 5,000 samples. A similar trend is noticed for shapicant as well. However, \textit{LLpowershap}, powershap and BorutaShap improve their performance with the additional samples. The reduced performance of \textit{LLpowershap} and powershap when 90\% of the 500 features are informative in the case of 5,000 samples compared to 10,000 samples (as shown in Table~\ref{make-classi-table1}) indicates the underfitting of the models due to limited number of samples to learn the underlying patterns.

\begin{table}[]
\caption{Simulation benchmark results for number of informative features returned on the datasets with 5,000 and 10,000 samples, created using \texttt{make\_classification} of \texttt{sklearn} with five different random seeds. The method with maximum number of highest informative features returned is shown in bold face.}
\label{make-classi-table1}
\begin{adjustbox}{margin=2cm 0.25cm 0cm 0cm}
\begin{tabular}{@{}lrrrrrrrr@{}}
\toprule
                                    & \multicolumn{4}{r@{}}{\textbf{\# Samples = 5,000}}  & \multicolumn{4}{r}{\textbf{\ \ \ \ \ \# Samples =10,000}}          \\ \midrule
\textbf{Method/ \# Feature}       & \textbf{\ \ 20} & \textbf{\ \ 100} & \textbf{\ \ 250} & \textbf{\ \ 500} & \textbf{\ \ \ \ \ 20} & \textbf{\ \ 100} & \textbf{\ \ \ 250} & \textbf{500} \\  \midrule
\textit{Informative features: 3\%} & \textbf{}   & \textbf{}    & \textbf{}    & \textbf{}    & \textbf{}   & \textbf{}    & \textbf{}    & \textbf{}    \\
\textbf{LLpowershap}                        & 1           & 3           & 7           & 15           & 1           & 3           & 7           & 15           \\
\textbf{powershap}                          & 1           & 3           & 7           & 15           & 1           & 3           & 7           & 15           \\
\textbf{BorutaShap}                          & 1           & 3           & 7           & 15            & 1           & 3           & 7           & 15           \\
\textbf{shapicant}                             & 1           & 3           & 7           & 15           & 1           & 3           & 7           & 15           \\ 
Chi²                                                & 1           & 1           & 5           & 9           & 1           & 1          & 5           & 9           \\
F Test                                            & 1           & 1           & 5           & 9            & 1           & 1           & 5           & 9          \\ \midrule

\textit{Informative features: 10\%} & \textbf{}   & \textbf{}    & \textbf{}    & \textbf{}    & \textbf{}   & \textbf{}    & \textbf{}    & \textbf{}    \\
\textbf{LLpowershap}                        & 2           & 10           & 25           & 50           & 2           & 10           & 25           & 50           \\
\textbf{powershap}                           & 2           & 10           & 25           & 50           & 2           & 10           & 25           & 50           \\
BorutaShap                                    & 2           & 10           & 23           & 34           & 2           & 10           & 25           & 47           \\
shapicant                                       & 2           & 10           & 23           & 42           & 2           & 10           & 25           & 49           \\ 
Chi²                                                & 1           & 6            & 15           & 28           & 1           & 6            & 15           & 28           \\
F Test                                             & 1           & 6            & 15           & 28           & 1           & 6            & 15           & 28           \\ \midrule

\textit{Informative features: 33\%} & \textbf{}   & \textbf{}    & \textbf{}    & \textbf{}    & \textbf{}   & \textbf{}    & \textbf{}    & \textbf{}    \\
\textbf{LLpowershap}         & 6           & 33           & 82           & 157          & 6           & 33           & 82           & 165          \\
powershap                           & 6           & 33           & 81           & 154          & 6           & 33           & 82           & 165          \\
BorutaShap                          & 6           & 31           & 51           & 67           & 6           & 33           & 69           & 104          \\
shapicant                           & 6           & 30           & 50           & 79           & 6           & 33           & 72           & 111          \\
Chi²                                & 4           & 21           & 52           & 106          & 4           & 21           & 52           & 105          \\
F Test                              & 4           & 21           & 52           & 100          & 4           & 21           & 52           & 103          \\  \midrule
\textit{Informative features: 50\%} & \textbf{}   & \textbf{}    & \textbf{}    & \textbf{}    & \textbf{}   & \textbf{}    & \textbf{}    & \textbf{}    \\
\textbf{LLpowershap}                         & 10          & 50           & 124          & 201          & 10          & 50           & 125          & 250          \\
powershap                           & 10          & 50           & 122          & 214          & 10          & 50           & 125          & 248          \\
BorutaShap                          & 10          & 37           & 70           & 66           & 10          & 48           & 95           & 140          \\
shapicant                           & 10          & 35           & 65           & 85           & 10          & 48           & 91           & 137          \\
Chi²                                & 6           & 28           & 82           & 165          & 6           & 28           & 83           & 168          \\
F Test                              & 6           & 28           & 81           & 147          & 6           & 28           & 81           & 160          \\  \midrule
\textit{Informative features: 90\%} & \textbf{}   & \textbf{}    & \textbf{}    & \textbf{}    & \textbf{}   & \textbf{}    & \textbf{}    & \textbf{}    \\
LLpowershap                         & 18          & 90           & 201          & 244          & 18          & 90           & 225          & 374          \\
\textbf{powershap}              & 18          & 90           & 202          & 260          & 18          & 90           & 225          & 394          \\
BorutaShap                          & 18          & 61           & 79           & 63           & 18          & 77           & 134          & 135          \\
shapicant                           & 18          & 48           & 67           & 81           & 18          & 69           & 121          & 149          \\
Chi²                                & 11          & 61           & 146          & 288          & 11          & 61           & 149          & 306          \\
F Test                              & 11          & 60           & 133          & 200          & 11          & 60           & 142          & 264          \\ \bottomrule
\end{tabular}
\end{adjustbox}
\end{table}

\begin{table}[]
\caption{Simulation benchmark results for number of noise features in the output on the datasets with 5,000 and 10,000 samples, created using \texttt{make\_classification} of \texttt{sklearn} with five different random seeds. Standard deviation values are given in brackets. Lowest noises are shown in bold face.}
\label{make-classi-table2}
\begin{tabular}{@{}lrrrrrrrr@{}} \\ \toprule
\textbf{}                    & \multicolumn{4}{c}{\textbf{\# Samples = 5,000}}                                                                                                                                                                               & \multicolumn{4}{c}{\textbf{\# Samples = 10,000}}                                                                                                                                                                              \\ \midrule
\textbf{Method\ /\ \# Feature} & \textbf{\ \ \ \ \ \ \ 20}                                           & \textbf{\ \ \ \ \ \ \ 100}                                          & \textbf{\ \ \ \ \ \ \ 250}                                          & \textbf{\ \ \ \ \ \ \ 500}                                          & \textbf{\ \ \ \ \ \ \ \ \ \ \ \ 20}                                           & \textbf{\ \ \ \ \ \ \ 100}                                          & \textbf{\ \ \ \ \ \ \ 250}                                          & \textbf{\ \ \ \ \ \ \ 500}                                          \\ \midrule
LLpowershap                  & \textbf{0(0)}                                                  & \begin{tabular}[c]{@{}r@{}}0.04\\ (0.20)\end{tabular} & \begin{tabular}[c]{@{}r@{}}0.36\\ (0.64)\end{tabular} &             \begin{tabular}[c]{@{}r@{}}0.88\\ (1.13)\end{tabular} & \textbf{0(0)}                                                  & \textbf{0(0)}                                                  & \begin{tabular}[c]{@{}r@{}}0.04\\ (0.20)\end{tabular} & \begin{tabular}[c]{@{}r@{}}0.36\\ (0.70)\end{tabular} \\[12pt]
powershap                    & \begin{tabular}[c]{@{}r@{}}0.12\\ (0.33)\end{tabular} & \begin{tabular}[c]{@{}r@{}}1.88\\ (1.83)\end{tabular} & \begin{tabular}[c]{@{}r@{}}2.44\\ (2.06)\end{tabular} & \begin{tabular}[c]{@{}r@{}}3.92\\ (3.99)\end{tabular}  & \begin{tabular}[c]{@{}r@{}}0.20\\ (0.41)\end{tabular} & \begin{tabular}[c]{@{}r@{}}0.88\\ (1.05)\end{tabular}  & \begin{tabular}[c]{@{}r@{}}2.36\\ (2.46)\end{tabular} & \begin{tabular}[c]{@{}r@{}}2.56\\ (2.75)\end{tabular} \\[10pt]

BorutaShap                   & \begin{tabular}[c]{@{}r@{}}0.16\\ (0.37)\end{tabular}  & \begin{tabular}[c]{@{}r@{}}0.16\\ (0.37)\end{tabular} & \begin{tabular}[c]{@{}r@{}}0.04\\ (0.20)\end{tabular}   & \begin{tabular}[c]{@{}r@{}}0.04\\ (0.20)\end{tabular} & \begin{tabular}[c]{@{}r@{}}0.08\\ (0.28)\end{tabular}  & \begin{tabular}[c]{@{}r@{}}0.08\\ (0.28)\end{tabular}  & \begin{tabular}[c]{@{}r@{}}0.12\\ (0.33)\end{tabular}  & \begin{tabular}[c]{@{}r@{}}0.08\\ (0.28)\end{tabular}  \\[10pt]
shapicant                    & \begin{tabular}[c]{@{}r@{}}1.12\\ (1.69)\end{tabular} & \begin{tabular}[c]{@{}r@{}}0.80\\ (1.58)\end{tabular} & \begin{tabular}[c]{@{}r@{}}0.64\\ (1.38)\end{tabular} & \begin{tabular}[c]{@{}r@{}}1.04\\ (1.70)\end{tabular} & \begin{tabular}[c]{@{}r@{}}5.84\\ (5.78)\end{tabular}  & \begin{tabular}[c]{@{}r@{}}2.92\\ (4.77)\end{tabular} & \begin{tabular}[c]{@{}r@{}}2.60\\ (4.37)\end{tabular} & \begin{tabular}[c]{@{}r@{}}2.36\\ (3.53)\end{tabular} \\[10pt]
Chi²                         & \textbf{0(0)}                                                  & \textbf{0(0)}                                                  & \textbf{0(0)}                                                  & \textbf{0(0)}                                                  & \textbf{0(0)}                                                  & \textbf{0(0)}                                           & \textbf{0(0)}                                                  & \textbf{0(0)}                                                  \\[8pt]
F Test                       & \begin{tabular}[c]{@{}r@{}}0.16\\ (0.47)\end{tabular} & \begin{tabular}[c]{@{}r@{}}0.64\\ (0.64)\end{tabular}  & \begin{tabular}[c]{@{}r@{}}1.40\\ (1.12)\end{tabular}  & \begin{tabular}[c]{@{}r@{}}2.76\\ (2.15)\end{tabular}  & \textbf{0(0)}                                                  & \begin{tabular}[c]{@{}r@{}}0.64\\ (0.70)\end{tabular}  & \begin{tabular}[c]{@{}r@{}}1.76\\ (1.85)\end{tabular} & \begin{tabular}[c]{@{}r@{}}3.12\\ (2.20)\end{tabular}  \\[10pt] \bottomrule
\end{tabular}
\end{table}

\subsection{Benchmark Datasets}

Table~\ref{selfeatures} shows the size of the selected feature sets on the benchmark classification datasets. As expected, the filter methods tend to select more features (except for F test in Madelon where the features have highly non-linear relationship with the outcome). BorutaShap picks the lowest number of features, probably due to the fact that selected features must be better than the best shadow feature. 

\begin{figure*}[t]
  \centering
  \includegraphics[width=\textwidth]{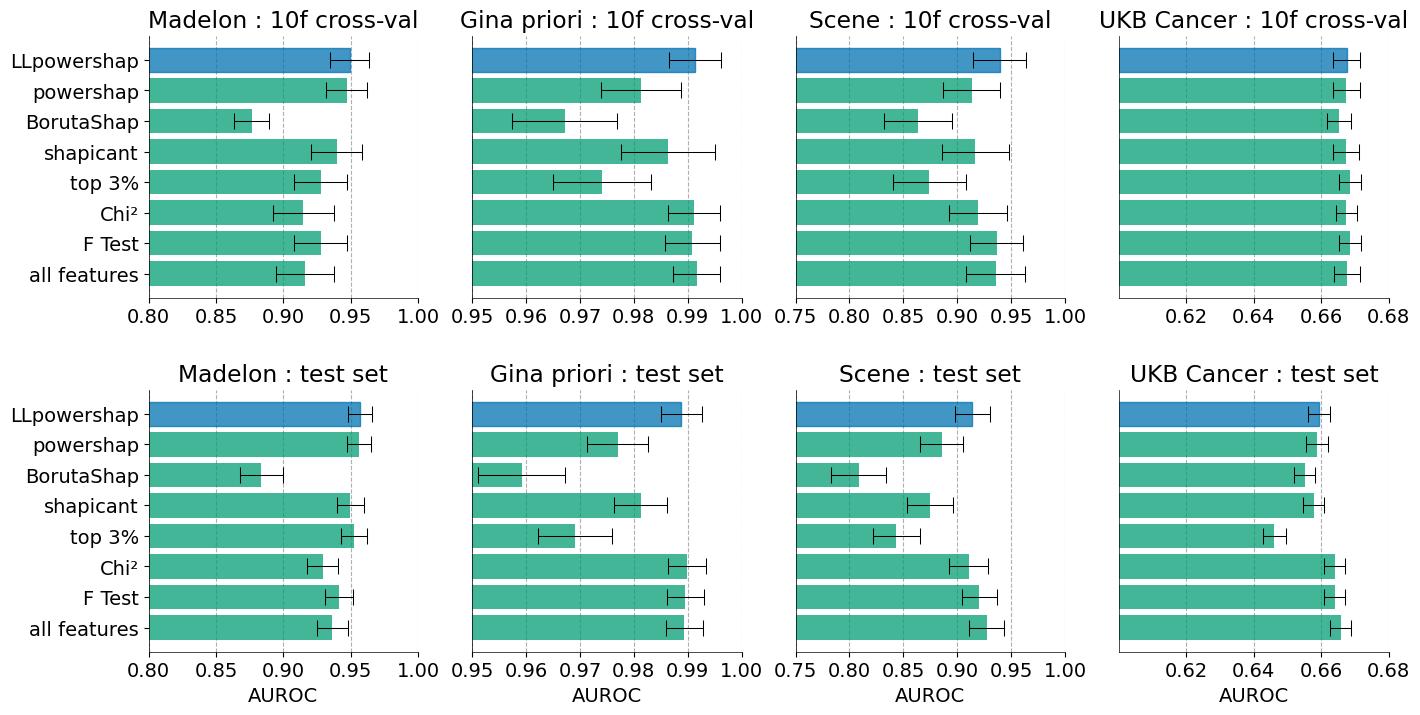}
  \caption{Benchmark performance using default CatBoost model, with error bars representing the standard deviations.}
  \label{fig3}
\end{figure*}

\begin{figure*}[t]
  \centering
  \includegraphics[width=\textwidth]{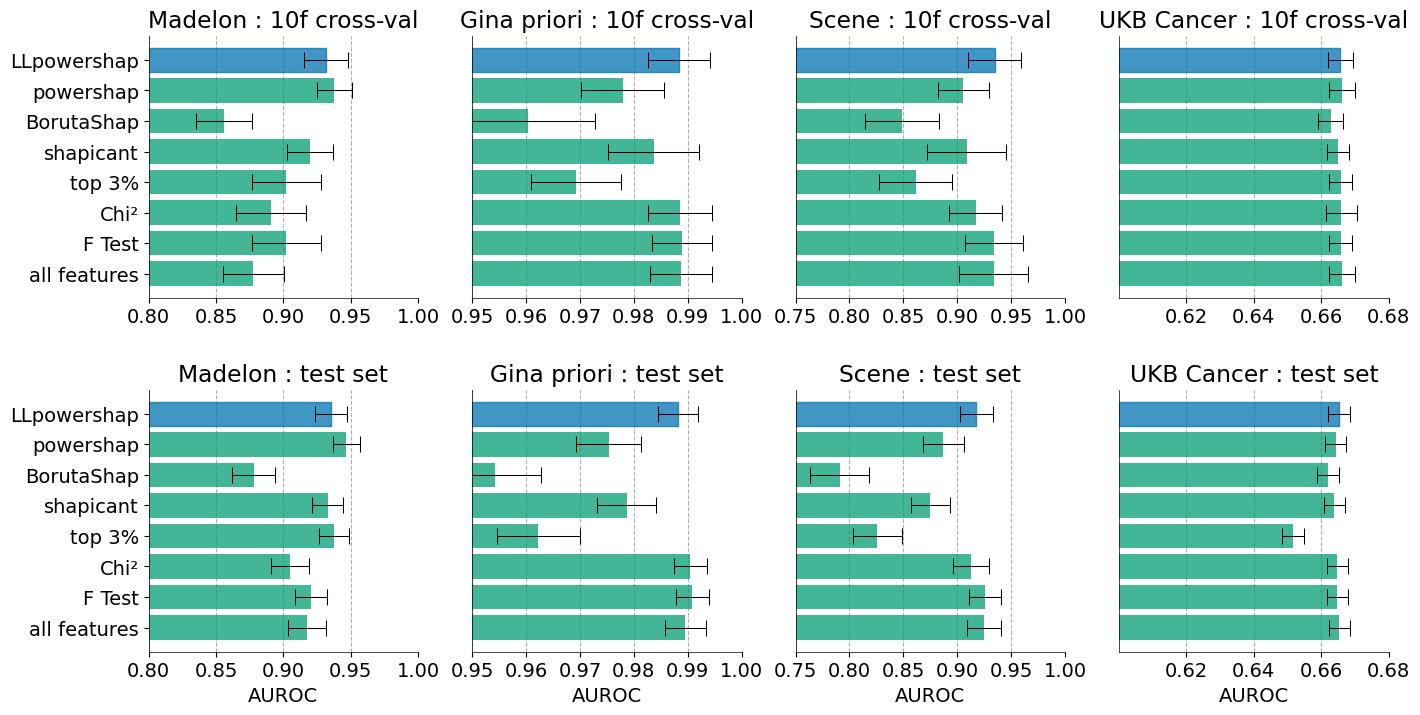}
  \caption{Benchmark performance using default LightGBM model, with error bars representing the standard deviations.}
  \label{fig4}
\end{figure*}

Performance of the selected features using 10-fold cross validation and on the test set is shown in Figure~\ref{fig3} for a default CatBoost model and in Figure~\ref{fig4} for a default LightGBM model, measured using AUROC. In most cases, \textit{LLpowershap} has the highest performance among the datasets and the methods. Other than the filter method F test (which needs additional feature creation to capture non-linearity and interaction), \textit{LLpowershap} is the only method achieving a performance with error bar overlapping that of the highest performance method in each scenario. The top 3\% method performs poorly on the Scene and Gina priori datasets but performs better on the Madelon dataset. For the UKB Cancer data, all methods perform well in the cross-validation. The top 3\% method performs poorly compared to the other methods on the test set. Interestingly, we see a very similar trend in performance (between Figure~\ref{fig3} and Figure~\ref{fig4}) when we use a CatBoost or a LightGBM to measure the performance of the selected features.

\begin{table}[h!]
  \centering
  \caption{Benchmark results for number of selected features.}
  \begin{tabular}{lrrrrrrrr}
    \hline
    \textbf{Dataset} & \textbf{LLpowershap} & \textbf{powershap} & \textbf{BorutaShap} & \textbf{shapicant} & \textbf{chi\(^2\)} & \textbf{F test} & \textbf{top 3\%} & \textbf{all features} \\ \midrule

    Madelon & 21 & 10 & 6 & 26 & 43 & 18 & 15 & 500 \\

    Gina priori & 123 & 43 & 15 & 41 & 628 & 405 & 23 & 784 \\

    Scene & 60 & 20 & 7 & 19 & 95 & 225 & 8 & 294 \\
    UKB Cancer & 41 & 30 & 18 & 117 & 578 & 589 & 84 & 2,228 \\
    \hline
  \end{tabular}
  \label{selfeatures}
\end{table}

\section{Discussion}\label{discussion}

Our experiments on simulated data as well as on the benchmark datasets demonstrate that there is advantage in using \textit{LLpowershap} in terms of selecting higher number of informative features with minimal or no noise, and also providing higher performance on the selected set of features. Further improvements could be achieved by replacing the default XGBoost models with hyper-parameter tuned XGBoost models and by optimising the wrapper hyper-parameters such as $\alpha$ and $\beta$ values. As seen with the simulation results for 500 features with 90\% informative features for 5,000 and 10,000 samples, resolving underfitting can be helpful in extracting even a higher number of informative features and for further reduction in noise features in the output. The other Shapley values based feature selection methods (especially shapicant and BorutaShap) may also benefit from using tuned models. Utilising powershap's convergence option, \textit{LLpowershap} can be used to extract maximum number of informative features by successively removing already found informative features from the input. Outputting lower number of noise features becomes highly relevant in this scenario. Unselected collinear features can also be recovered by using the convergence option. Benchmark results on the dataset Gina priori show the weakness of filter methods. Chi-squared method selects 80\% of the features as important and F test selects 52\% of the features as important. Although \textit{LLpowershap} selects only 16\% of the features as important, it has equal performance with the filter methods in 10-fold cross validation using both CatBoost and LightGBM models.

\textit{LLpowershap} tends to be slower than powershap as time complexity of calculating Shapley values using Interventional TreeSHAP scales in terms of model size and the size of the background dataset. We set background dataset to the maximum recommended size of 1024 samples. Path-dependent TreeSHAP (as employed by powershap, BorutaShap and shapicant) does not suffer from this issue. In the absence of this issue, both \textit{LLpowershap} and powershap should have similar time complexity. However, on the Madelon dataset, we find \textit{LLpowershap} is faster than BorutaShap and shapicant but slower than powershap. On the Gina priori and Scene datasets, \textit{LLpowershap} is faster than shapicant  but slower than BorutaShap. On the UKB Cancer dataset with over 450,000 samples and over 2,800 features, among the wrapper methods, powershap was the fastest, followed by LLpowershap, the top 3\% method, BorutaShap and shapicant.

Every study comes with some limitations and this one is no exception. As noted in~\cite{verhaeghe2022powershap}, \texttt{make\_classification} in its  default mode generates data points from hypercubes as linear classification problems, which are easier to classify. Also, as the no-free lunch theorems state that there is no one single algorithm exists, that performs best on all possible datasets, there is still the need to tune hyper-parameters of models used within a wrapper and to optimise the hype-parameters of the wrapper method as well for specific datasets. The results of such time consuming actions could be different from what we have  presented here, but these analyses suggest that there are inherent benefits in using loss-based Shapley values on unseen data. 

Before opting for five noise features of different distributions, we experimented with the idea of permuting the most important feature (by Shapley values) as the single noise feature replacing the uniform random noise in powershap but our experiments showed that that this may not beneficial, at least on the simulation data in this work. We also tried using uniform noise with very small correlation (<1e-6) with the outcome variable and our experiments showed this may not be particularly useful, given such noise features might have higher correlation with other features, thereby becoming useful in the construction of decision trees. We also experimented with loss-based Shapley values on the training and validation sets and our experiments showed that they are not as effective as Shapley values on the test set.

For rank-based feature selection methods, it is difficult to set a threshold that works for many datasets for discarding irrelevant features. Different strategies include defining a threshold by the largest gap between two consecutively ranked features~\cite{sanchez2007filter} or heuristically considering anything below the top 3\% of features (if there are more than hundred features to select from) after ranking as irrelevant, as used in~\cite{bolon2015feature}.  From a predictive performance perspective, our tests on the benchmark datasets showed this strategy effective only in the case of the dataset Madelon and to some extent on UKB Cancer dataset, advocating the need to check the performance of the models with the selected features before conducting further follow-up analyses. Alternatively, methods such as \textit{LLpowershap} outputting features based on given cut-off value for p-values are more effective. 

With the advent of big data, we have the opportunity  to deal with large scale datasets (both in terms of samples and features), making it more important to do feature selection as an initial screening step in the analysis pipeline. Often these large datasets contain a very high proportion of features that do not have any relevance to the outcome of interest. For example, large scale biomedical datasets containing comprehensive information (including hospital admissions, treatment and medication history) on participants used for discovering risk factors of rare diseases. Our simulation results with only 3\% of the total number of features as important when there are sufficiently large number of features (250 and 500) show a miniature of this scenario. \textit{LLpowershap}'s better performance in terms of noise output in such a scenario (for example, 3\% informative features out of a total of 500 features, bottom right plot in Figure~\ref{fig1}) may suggest that our method is more suitable for large scale real-world datasets with a very small proportion of relevant features with respect to the outcome of interest.

\section{Conclusion}\label{conclusion}

We propose \textit{LLpowershap} (logistic loss-based powershap) as a viable wrapper feature selection method for classification on a wide range of datasets as our experiments show that they can output increased number of informative features with lower number of noise features in the output. On the benchmark real-world datasets, our method shows consistent higher or at par performance on the selected features, tested using two independent state-of-the-art gradient boosting decision tree algorithms.

\subsubsection{Acknowledgements.} This project was funded by Medical Research Future Fund (MRFF), Grant/Award Number: 2007431. Elina Hypp\"{o}nen is funded by NHMRC Investigator Fellowship (GT2025349).

We acknowledge the source code made available by the authors of the package powershap available at \url{https://github.com/predict-idlab/powershap}, which were selectively modified by us to create \textit{LLpowershap}. The code for \textit{LLpowershap} can be found at \url{https://github.com/madakkmi/LLpowershap}. 

This research has been conducted using the UK Biobank Resource under Application Number 89630. The UK Biobank is an open access resource and bona fide researchers can apply to use the UK Biobank dataset by registering and applying at \url{https://www.ukbiobank.ac.uk/enable-yourresearch/register}.

\subsubsection{Contributions.}  Elina Hypp\"{o}nen and Iqbal Madakkatel conceived the idea. Iqbal Madakkatel designed and programmed the methodology and drafted the manuscript. Elina Hypp\"{o}nen managed funding acquisition, supervised, reviewed and edited the manuscript.

\subsubsection{Conflicts of interest.} The authors declare that there are no competing interests.


\renewcommand{\bibname}{References}
\bibliographystyle{unsrtnat}
 \bibliography{llpowershap}

\end{document}